\documentclass[11pt,a4paper]{article}
\usepackage{acl2001,times}

\makeatletter
\renewcommand{\normalsize}{\@setsize\normalsize{12.5pt}\xipt\@xipt} 
\makeatother

\usepackage{graphics}
\usepackage{epic}
\usepackage{amssymb}
\usepackage{pstricks}
\usepackage{pst-node}

\newcommand{\order}[1]{${\mathcal{O}}(#1)$}

\newcommand{\tabbox}[2]
 { \rnode{#1}{\psframebox{\begin{tabular}{@{\hspace{1mm}}c@{\hspace{1mm}}} #2 \end{tabular}}} }

\title { Computational properties of 
         environment-based disambiguation \footnotemark[1] }

\author
 {
   William Schuler \\
   Department of Computer and Information Science \\
   University of Pennsylvania \\
   Philadelphia, PA 19103 \\
   {\tt schuler@linc.cis.upenn.edu} \\
 }

\begin{document}
 
\maketitle

\def\thefootnote{\fnsymbol{footnote}}
\footnotetext[1]
 {
 The author would like to thank the other members of the Actionary research group, 
 Jan Allbeck, Koji Ashida, Norm Badler, Aravind Joshi, Mike Johns, Karin Kipper, 
 Mike Moore, and Martha Palmer, and the anonymous reviewers for their valuable comments.
 This work was partially supported by NSF CISE IIS99-00297 and ARO Grant
 DAAH0404-94-GE-0426.
 }
\def\thefootnote{\arabic{footnote}}

\begin{abstract}
 { The standard pipeline approach to semantic processing, in which sentences are 
   morphologically and syntactically resolved to a single tree before they are 
   interpreted, is a poor fit for applications such as natural language interfaces. 
   This is because the environment information, in the form of the objects and events 
   in the application's run-time environment, cannot be used to inform parsing decisions 
   unless the input sentence is semantically analyzed, but this does not occur 
   until after parsing in the single-tree semantic architecture.
   This paper describes the computational properties of an alternative architecture, 
   in which semantic analysis is performed on all possible interpretations 
   {\em during} parsing, in polynomial time.
 }
\end{abstract}


\section{Introduction}

Shallow semantic processing applications, comparing argument 
structures to search patterns or filling in simple templates, can 
achieve respectable results using the standard `pipeline' approach to 
semantics, in which sentences are morphologically and syntactically 
resolved to a single tree before being interpreted.
Putting disambiguation ahead of semantic evaluation is reasonable in 
these applications because they are primarily run on content 
like newspaper text or dictated speech, where no machine-readable 
contextual information is readily available to provide semantic 
guidance for disambiguation.

This single-tree semantic architecture is a poor fit for applications such 
as natural language interfaces however, in which a large amount of contextual 
information is available in the form of the objects and events in the 
application's run-time environment.
This is because the environment information cannot be used to inform parsing and 
disambiguation decisions unless the input sentence is semantically analyzed, 
but this does not occur until after parsing in the single-tree architecture.
Assuming that no current statistical disambiguation technique is 
so accurate that it could not benefit from this kind of environment-based 
information (if available), then it is important that the semantic 
analysis in an interface architecture be efficiently performed {\em during} 
parsing.

This paper describes the computational properties of one such architecture, 
embedded within a system for giving various kinds of conditional instructions 
and behavioral constraints to virtual human agents in a 3-D simulated environment 
\cite{agents00}.
In one application of this system, users direct simulated maintenance 
personnel to repair a jet engine, in order to ensure that the maintenance 
procedures do not risk the safety of the people performing them.
Since it is expected to process a broad range of maintenance instructions, 
the parser is run on a large subset of the Xtag English grammar \cite{xtag98}, 
which has been annotated with lexical semantic classes \cite{kipperetal00} 
associated with the objects, states, and processes in the maintenance simulation.
Since the grammar has several thousand lexical entries, the parser is exposed 
to considerable lexical and structural ambiguity as a matter of course.

The environment-based disambiguation architecture described in this paper 
has much in common with very early environment-based 
approaches, such as those described by Winograd \cite{winograd72},
in that it uses the actual entities in an environment database to resolve 
ambiguity in the input.
This research explores two extensions to the basic approach however:
\begin{enumerate}
  \item It incorporates ideas from type theory to represent a broad range  
        of linguistic phenomena in a manner for which their extensions 
        or {\em potential referents} in the environment are well-defined 
        in every case.
        This is elaborated in Section~\ref{sect:nodes}.
  \item It adapts the concept of structure sharing, taken from the study of 
        parsing, not only to translate the many possible interpretations of 
        ambiguous sentences into shared logical expressions, but also to 
        evaluate these sets of potential referents, over all possible 
        interpretations, in polynomial time.
        This is elaborated in Section~\ref{sect:forest}.
\end{enumerate}
Taken together, these extensions allow interfaced systems to 
evaluate a broad range of natural language inputs -- including those 
containing NP/VP attachment ambiguity and verb sense ambiguity -- 
in a principled way, simply based on the objects and events in the 
systems' environments.
For example, such a system would be able to correctly answer 
`Did someone stop the test at 3:00?' and resolve the ambiguity 
in the attachment of `at 3:00' 
just from the fact that there aren't any 3:00 
tests in the environment, only an event where one stops at 3:00.%
\footnote{It is important to make a distinction between this 
{\em environment information}, which just describes the set of 
objects and events that exist in the interfaced application, 
and what is often called {\em domain information}, which describes 
(usually via hand-written rules) the kinds of objects and events 
{\em can} exist in the interfaced application.
The former comes for free with the application, while the latter 
can be very expensive to create and port between domains.}
Because it evaluates instructions before attempting to choose a single 
interpretation, the interpreter can avoid getting `stranded' by 
disambiguation errors in earlier phases of analysis.

The main challenge of this approach is that it requires the efficient calculation 
of the set of objects, states, or processes in the environment that each possible 
sub-derivation of an input sentence could refer to.
A semantic interpreter could always be run on an (exponential) enumerated set 
of possible parse trees as a post-process, to filter out those interpretations 
which have no environment referents, but recomputing the potential environment 
referents for every tree would require an enormous amount of time (particularly 
for broad coverage grammars such as the one employed here).
The primary result of this paper is therefore a method of containing the time 
complexity of these calculations to lie within the complexity of parsing 
(i.e.~within \order{n^3} for a context-free grammar, where $n$ is the number 
of words in the input sentence), without sacrificing logical correctness, 
in order to make environment-based interpretation tractable for interactive 
applications.

\section{Representation of referents}
\label{sect:nodes}

Existing environment-based methods (such as those proposed by Winograd) 
only calculate the referents of noun phrases, so they only consult the 
{\em objects} in an environment database when interpreting input sentences.
But the evaluation of ambiguous sentences will be incomplete if the 
referents for verb phrases and other predicates are not calculated. 
In order to evaluate the possible interpretations of a sentence, 
as described in the previous section, an interface needs to define 
referent sets for every possible constituent.%
\footnote{This is not strictly true, as 
referent sets for constituents like determiners are difficult to define, 
and others (particularly those of quantifiers) will be extremely large 
until composed with modifiers and arguments.
Fortunately, as long as there is a bound on the height in the tree to which the 
evaluation of referent sets can be deferred (e.g.~after the first composition), 
the claimed polynomial complexity of referent annotation will not be lost.}

The proposed solution draws on a theory of constituent types from 
formal linguistic semantics, in which constituents such as nouns and 
verb phrases are represented as composeable functions that take entitiess 
or situations as inputs and ultimately return a truth value for the sentence.
Following a straightforward adaptation of standard type theory, common 
nouns (functions from entities to truth values) define potential referent 
sets of simple environment entities: $\{e_1,e_2,e_3,\dots\}$, and sentences 
(functions from situations or world states to truth values) define 
potential referent sets of situations in which those sentences hold true: 
$\{\sigma_1,\sigma_2,\sigma_3,\dots\}$.
Depending on the needs of the application, these situations can be represented 
as intervals along a time line \cite{allenferguson94}, or as regions in a 
three-dimensional space \cite{xubadler00}, or as some combination of the two, 
so that they can be constrained by modifiers that specify the situations' times 
and locations.
Referents for other types of phrases may be expressed as tuples of entities 
and situations: one for each argument of the corresponding logical function's 
input (with the presence or absence of the tuple representing the boolean output). 
For example, adjectives, prepositional phrases, and relative clauses, which are 
typically represented as situationally-dependent properties (functions from 
situations and entities to truth values) define potential referent sets of tuples 
that consist of one entity and one situation: $\{\langle e_1,\sigma_1 \rangle, 
\langle e_2,\sigma_2 \rangle, \langle e_3,\sigma_3 \rangle, \dots\}$.
This representation can be extended to treat common nouns as situationally-dependent 
properties as well, in order to handle sets like `bachelors' that 
change their membership over time.

\section{Sharing referents across interpretations}
\label{sect:forest}

Any method for using the environment to guide the interpretation of natural 
language sentences requires a tractable representation of the many possible 
interpretations of each input.
The representation described here is based on the polynomial-sized 
chart produced by any dynamic programming recognition algorithm.

\begin{figure*}[htbp]

\begin{center}
\resizebox{.8\linewidth}{!}{
\begin{pspicture}(16,11)
\rput(2,1) { \tabbox{p1}{ NP[button] \\ 
                          $button'(x)$ \\
                          $\{b_1,b_2,b_3\}$ } }
\rput(5,1) { \tabbox{p2}{ P[on] \\ 
                          $on'(x,y)$ \\
                          $\!\!\!\{\langle b_1,h_1 \rangle, \langle h_2,d_1 \rangle\}\!\!\!\!$ } }
\rput(8,1) { \tabbox{p3}{ NP[handle] \\ 
                          $handle'(x)$ \\
                          $\{h_1,h_2,h_3\}$ } }
\rput(11,1) { \tabbox{p4}{ P[beside] \\ 
                           $beside'(x,y)$ \\
                           $\!\!\!\{\langle b_1,a_1 \rangle, \langle r_1,a_1 \rangle\}\!\!\!\!$ } }
\rput(14,1) { \tabbox{p5}{ NP[adapter] \\ 
                           $adapter'(x)$ \\
                           $\{a_1,a_2,a_3\}$ } }
\rput(6.5,3.5) { \tabbox{p23}{ PP[on] \\  $on''(x,y)$ \\ $\{\langle b_1,h_1 \rangle\}$ } }
\rput(12.5,3.5) { \tabbox{p45}{ PP[beside] \\  $beside''(x,y)$ \\ $\{\langle b_1,a_1 \rangle, \langle r_1,a_1 \rangle\}$ } }
\rput(5,6) { \tabbox{p13}{ NP[button] \\ $button'(x)$ \\ $\{b_1\}$ } }
\rput(11,6) { \tabbox{p35}{ NP[handle] \\  $handle'(x)$ \\ $\emptyset$ } }
\rput(9.5,8.5) { \tabbox{p25}{ PP[on] \\ $on''(x,y)$ \\ $\emptyset$ } }
\rput(8,11) { \tabbox{p15}{ NP[button] \\ $button'(x)$ \\ $\{b_1\}$ } }
\cnode(6.5,2.25){2mm}{p23a}
  \ncline{p2}{p23a} \ncline{p3}{p23a}
  \ncline{p23a}{p23}
\cnode(12.5,2.25){2mm}{p45a}
  \ncline{p4}{p45a} \ncline{p5}{p45a}
  \ncline{p45a}{p45}
\cnode(5,4.75){2mm}{p13a}
  \ncline{p1}{p13a} \ncline{p23}{p13a}
\cnode(11,4.75){2mm}{p35a}
  \ncline{p3}{p35a} \ncline{p45}{p35a}
  \ncline{p13a}{p13}
  \ncline{p35a}{p35}
\cnode(9.5,7.25){2mm}{p25a}
  \nccurve[angleA=105,angleB=200]{p2}{p25a} \ncline{p35}{p25a}
  \ncline{p25a}{p25}
\cnode(8.5,9.75){2mm}{p15a}
  \ncline{p13}{p15a} \nccurve[angleA=60,angleB=0]{p45}{p15a}
\cnode(7.5,9.75){2mm}{p15b}
  \nccurve[angleA=120,angleB=180]{p1}{p15b} \ncline{p25}{p15b}
  \ncline{p15a}{p15} \ncline{p15b}{p15}
\end{pspicture}
}
\end{center}

\caption{Example shared forest for ``button on handle beside adapter''}
\label{fig:gateforest}

\end{figure*}
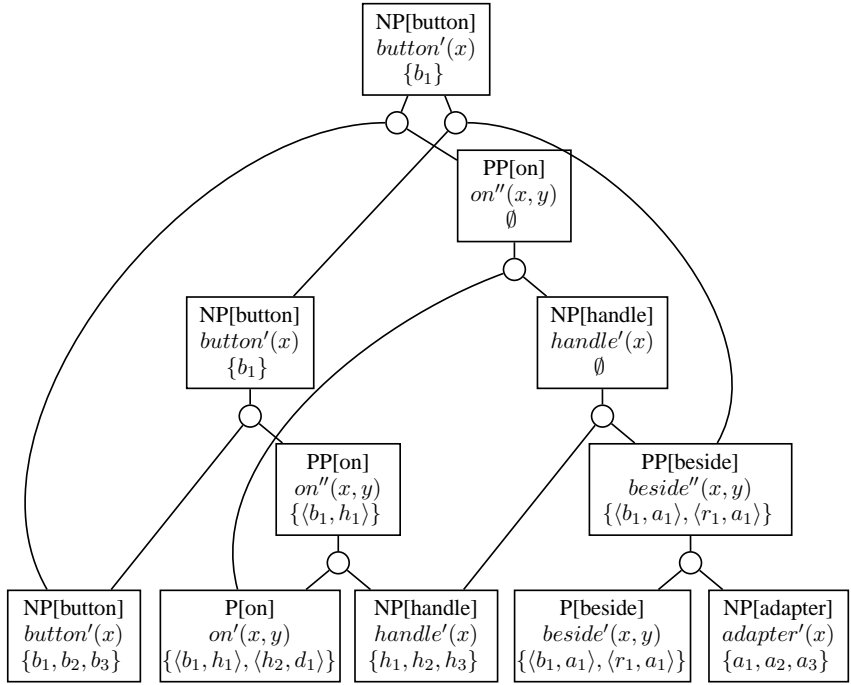

A record of the derivation paths in any dynamic programming recognition 
algorithm (such as CKY \cite{cockeschwartz70,kasami65,younger67} or 
Earley \cite{earley70}) can be interpreted as a polynomial sized and-or graph 
with space complexity equal to the time complexity of recognition, 
whose disjunctive nodes represent possible constituents in the analysis,
and whose conjunctive nodes represent binary applications of rules in the grammar. 
This is called a {\em shared forest} of parse trees, because it can represent 
an exponential number of possible parses using a polynomial number of nodes 
which are shared between alternative analyses \cite{tomita85,billotlang89}, and 
can be constructed and traversed in time of the same complexity 
(e.g.~\order{n^3} for context free grammars).
For example, the two parse trees for the noun phrase 
`button on handle beside adapter' shown in Figure~\ref{fig:gatetrees} 
can be merged into the single shared forest in Figure~\ref{fig:gateforest}
without any loss of information.


\begin{figure*}[htbp]

\begin{center}
\resizebox{.8\linewidth}{!}{
\begin{pspicture}(16,12)
\rput(2,1) { \tabbox{p1}{ NP[button] \\ 
                          $button'(x)$ \\
                          $\{b_1,b_2,b_3\}$ } }
\rput(5,1) { \tabbox{p2}{ P[on] \\ 
                          $on'(x,y)$ \\
                          $\!\!\!\{\langle b_1,h_1 \rangle, \langle h_2,d_1 \rangle\}\!\!\!\!$ } }
\rput(8,1) { \tabbox{p3}{ NP[handle] \\ 
                          $handle'(x)$ \\
                          $\{h_1,h_2,h_3\}$ } }
\rput(11,1) { \tabbox{p4}{ P[beside] \\ 
                           $beside'(x,y)$ \\
                           $\!\!\!\{\langle b_1,a_1 \rangle, \langle r_1,a_1 \rangle\}\!\!\!\!$ } }
\rput(14,1) { \tabbox{p5}{ NP[adapter] \\ 
                           $adapter'(x)$ \\
                           $\{a_1,a_2,a_3\}$ } }
\rput(12.5,3.5) { \tabbox{p45}{ PP[beside] \\  $beside''(x,y)$ \\ $\{\langle b_1,a_1 \rangle, \langle r_1,a_1 \rangle\}$ } }
  \ncline{p4}{p45} \ncline{p5}{p45}
\rput(11,6) { \tabbox{p35}{ NP[handle] \\  $handle'(x)$ \\ $\emptyset$ } }
  \ncline{p3}{p35} \ncline{p45}{p35}
\rput(9.5,8.5) { \tabbox{p25}{ PP[on] \\ $on''(x,y)$ \\ $\emptyset$ } }
  \ncline{p2}{p25} \ncline{p35}{p25}
\rput(8,11) { \tabbox{p15}{ NP[button] \\ $button'(x)$ \\ $\emptyset$ } }
  \ncline{p1}{p15} \ncline{p25}{p15}
\end{pspicture}
}

\vspace{5mm}

\resizebox{.8\linewidth}{!}{
\begin{pspicture}(16,12)
\rput(2,1) { \tabbox{p1}{ NP[button] \\ 
                          $button'(x)$ \\
                          $\{b_1,b_2,b_3\}$ } }
\rput(5,1) { \tabbox{p2}{ P[on] \\ 
                          $on'(x,y)$ \\
                          $\!\!\!\{\langle b_1,h_1 \rangle, \langle h_2,d_1 \rangle\}\!\!\!\!$ } }
\rput(8,1) { \tabbox{p3}{ NP[handle] \\ 
                          $handle'(x)$ \\
                          $\{h_1,h_2,h_3\}$ } }
\rput(11,1) { \tabbox{p4}{ P[beside] \\ 
                           $beside'(x,y)$ \\
                           $\!\!\!\{\langle b_1,a_1 \rangle, \langle r_1,a_1 \rangle\}\!\!\!\!$ } }
\rput(14,1) { \tabbox{p5}{ NP[adapter] \\ 
                           $adapter'(x)$ \\
                           $\{a_1,a_2,a_3\}$ } }
\rput(12.5,3.5) { \tabbox{p45}{ PP[beside] \\  $beside''(x,y)$ \\ $\{\langle b_1,a_1 \rangle, \langle r_1,a_1 \rangle\}$ } }
  \ncline{p4}{p45} \ncline{p5}{p45}
\rput(6.5,3.5) { \tabbox{p23}{ PP[on] \\  $on''(x,y)$ \\ $\{\langle b_1,h_1 \rangle\}$ } }
  \ncline{p2}{p23} \ncline{p3}{p23}
\rput(5,6) { \tabbox{p13}{ NP[button] \\ $button'(x)$ \\ $\{b_1\}$ } }
  \ncline{p1}{p13} \ncline{p23}{p13}
\rput(8,11) { \tabbox{p15}{ NP[button] \\ $button'(x)$ \\ $\{b_1\}$ } }
  \ncline{p13}{p15} \ncline{p45}{p15}
\end{pspicture}
}
\end{center}

\caption{Example parse trees for `button on handle beside adapter'}
\label{fig:gatetrees}

\end{figure*}

These shared syntactic structures can further be associated with compositional 
semantic functions that correspond to the syntactic elements in the forest, to 
create a shared forest of trees each representing a complete expression in 
some logical form.
This extended sharing is similar to the `packing' approach employed in the 
Core Language Engine \cite{alshawi92}, except that the CLE relies on a quasi-logical 
form to underspecify semantic information such as quantifier scope 
(the calculation of which is deferred until syntactic ambiguities have been at 
least partially resolved by other means);
whereas the approach described here extends structure 
sharing to incorporate a certain amount of quantifier scope ambiguity in order 
to allow a complete evaluation of all subderivations in a shared forest before 
making any disambiguation decisions in syntax.%
\footnote{A similar basis on (at least partially) disambiguated syntactic 
representations makes similar underspecified semantic representations such as 
hole semantics \cite{bos95} ill-suited for environment-based syntactic disambiguation.}
Various synchronous formalisms have been introduced for associating syntactic 
representations with logical functions in isomorphic or locally non-isomorphic 
derivations, including Categorial Grammars (CGs) \cite{woods93}, 
Synchronous Tree Adjoining Grammars (TAGs) \cite{joshi85,shieberschabes90,shieber94}, 
and Synchronous Description Tree Grammars (DTGs) \cite{rambowetal95,rambowsatta96}.
Most of these formalisms can be extended to define semantic associations over 
entire shared forests, rather than merely over individual parse trees, in a 
straightforward manner, preserving the ambiguity of the syntactic forest without 
exceeding its polynomial size, or the polynomial time complexity of creating or 
traversing it.

Since one of the goals of this architecture is to use the system's representation of 
its environment to resolve ambiguity in its instructions, a space-efficient 
shared forest of logical functions will not be enough.
The system must also be able to efficiently calculate the sets of potential 
referents in the environment for every subexpression in this forest.
Fortunately, since the logical function forest shares structure between alternative 
analyses, many of the sets of potential referents can be shared between 
analyses during evaluation as well.
This has the effect of building a third shared forest of potential referent sets 
(another and-or graph, isomorphic to the logical function forest and with the same 
polynomial complexity), where every conjunctive node represents the results of 
applying a logical function to the elements in that node's child sets, 
and every disjunctive node represents the union of all the 
potential referents in that node's child sets.
The presence or absence of these environment referents at various nodes in the 
shared forest can be used to choose a viable parse tree from the forest, 
or to evaluate the truth or falsity of the input sentence without disambiguating it 
(by checking the presence or lack of referents at the root of the forest).

For example, the noun phrase `button on handle beside adapter' 
has at least two possible interpretations, represented by the two trees 
in Figure~\ref{fig:gatetrees}:
one in which a button is on a handle and the {\em handle} 
(but not necessarily the button) is beside an adapter, 
and the other in which a button is on a handle and the {\em button} 
(but not necessarily the handle) is beside an adapter.
The semantic functions are annotated just below the syntactic categories, and the 
potential environment referents are annotated just below the semantic functions 
in the figure.
Because there are no handles next to adapters in the environment (only buttons 
next to adapters), the first interpretation has no environment referents at its 
root, so this analysis is dispreferred if it occurs within the analysis of a 
larger sentence.
The second interpretation does have potential environment referents all the way 
up to the root (there is a button on a handle which is also beside an adapter), 
so this analysis is preferred if it occurs within the analysis of a larger sentence.

The shared forest representation effectively merges the enumerated set 
of parse trees into a single data structure, and unions the referent sets 
of the nodes in these trees that have the same label and cover the same span 
in the string yield (such as the root node, leaves, and the PP covering 
`beside adapter' in the examples above).
The referent-annotated forest for this sentence therefore looks like 
the forest in Figure~\ref{fig:gateforest}, in which the sets of buttons, 
handles, and adapters, as well as the set of things beside adapters, 
are shared between the two alternative interpretations.
If there is a button next to an adapter, but no handle next to an adapter, 
the tree representing `handle beside adapter' as a constituent 
may be dispreferred in disambiguation, but the NP constituent at the root 
is still preferred because it has potential referents in the environment 
due to the other interpretation.

\begin{figure*}[htbp]

\begin{center}
\resizebox{.8\linewidth}{!}{
\begin{pspicture}(16,11)
\rput(2,1) { \tabbox{p1}{ VP[drained] \\ 
                          $drain'(x)$ \\
                          $\{\langle \sigma_1, e_1 \rangle, \langle \sigma_2, e_2 \rangle\} \!\!$ } }
\rput(5,1) { \tabbox{p2}{ P[after] \\ 
                          $after'(x,y)$ \\
                          \dots } }
\rput(8,1) { \tabbox{p3}{ NP[test] \\ 
                          $test'(x)$ \\
                          $\{\tau_1,\tau_2,\tau_3\}$ } }
\rput(11,1) { \tabbox{p4}{ P[at] \\ 
                           $at'(x,y)$ \\
                           \dots } }
\rput(14,1) { \tabbox{p5}{ NP[3:00] \\ 
                           constant $t_1$ \\
                           $\{t_1\}$ } }
\rput(6.5,3.5) { \tabbox{p23}{ PP[after] \\  $after''(x,y)$ \\ $\{\langle \sigma_1,\tau_1 \rangle\}$ } }
\rput(12.5,3.5) { \tabbox{p45}{ PP[at] \\  $at''(x,y)$ \\ $\{\langle \sigma_1,t_1 \rangle, \langle \rho_1,t_1 \rangle\}$ } }
\rput(5,6) { \tabbox{p13}{ VP[drained] \\ $drain'(x)$ \\ $\{\langle \sigma_1, e_1 \rangle\} \!\!$ } }
\rput(11,6) { \tabbox{p35}{ NP[test] \\  $test'(x)$ \\ $\emptyset$ } }
\rput(9.5,8.5) { \tabbox{p25}{ PP[after] \\ $after''(x,y)$ \\ $\emptyset$ } }
\rput(8,11) { \tabbox{p15}{ VP[drained] \\ $drain'(x)$ \\ $\{\langle \sigma_1, e_1 \rangle\} \!\!$ } }
\cnode(6.5,2.25){2mm}{p23a}
  \ncline{p2}{p23a} \ncline{p3}{p23a}
  \ncline{p23a}{p23}
\cnode(12.5,2.25){2mm}{p45a}
  \ncline{p4}{p45a} \ncline{p5}{p45a}
  \ncline{p45a}{p45}
\cnode(5,4.75){2mm}{p13a}
  \ncline{p1}{p13a} \ncline{p23}{p13a}
\cnode(11,4.75){2mm}{p35a}
  \ncline{p3}{p35a} \ncline{p45}{p35a}
  \ncline{p13a}{p13}
  \ncline{p35a}{p35}
\cnode(9.5,7.25){2mm}{p25a}
  \nccurve[angleA=105,angleB=200]{p2}{p25a} \ncline{p35}{p25a}
  \ncline{p25a}{p25}
\cnode(8.5,9.75){2mm}{p15a}
  \ncline{p13}{p15a} \nccurve[angleA=60,angleB=0]{p45}{p15a}
\cnode(7.5,9.75){2mm}{p15b}
  \nccurve[angleA=120,angleB=180]{p1}{p15b} \ncline{p25}{p15b}
  \ncline{p15a}{p15} \ncline{p15b}{p15}
\end{pspicture}
}
\end{center}

\caption{Example shared forest for verb phrase ``drained after test at 3:00''}
\label{fig:testforest}

\end{figure*}
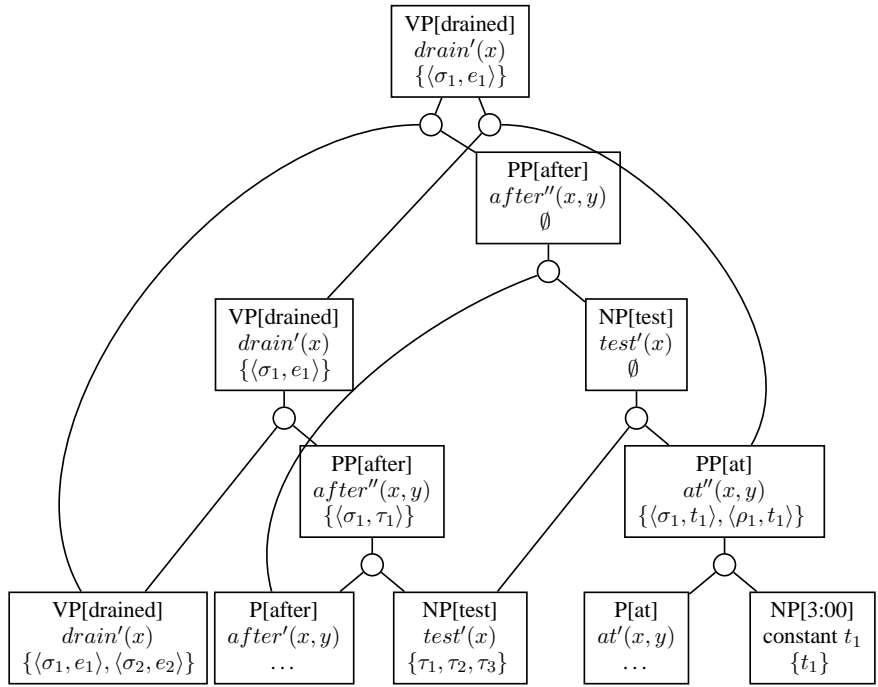

The logical function at each node is defined over the referent sets of that 
node's immediate children.
Nodes that represent the attachment of a modifier with referent set $M$ to a 
relation with referent set $R$ produce referent sets of the form: 

\vspace{3mm} \( 
\{ \ \langle x,\dots \rangle \ | 
   \ \langle x,\dots \rangle \!\in\! R \ \wedge
   \ \langle x \rangle \!\in\! M \ \} 
\) \vspace{3mm}

\noindent Nodes in a logical function forest that represent the attachment of an 
argument with referent set $A$ to a relation with referent set $R$ produce 
referent sets of the form: 

\vspace{3mm} \( 
\{ \ \langle \dots \rangle \ | 
   \ \exists x . \langle \dots,x \rangle \!\in\! R \ \wedge
   \ \langle x \rangle \!\in\! A \ \} 
\) \vspace{3mm}

\noindent effectively stripping off one of the objects in each tuple 
if the object is also found in the set of referents for the argument.%
\footnote{In order to show where the referents came from, the tuple objects 
are not stripped off in Figures~\ref{fig:gatetrees} and~\ref{fig:gateforest}.
Instead, an additional bar is added to the function name to designate the 
effective last object in each tuple: 
the tuple $\langle b_1,a_1 \rangle$ referenced by $beside'$ has $a_1$ as the 
last element, but the tuple referenced by $beside''$ actually has $b_1$ as 
the last element since the complement $a_1$ has been already been attached.}
This is a direct application of standard type theory to the calculation 
of referent sets: modifiers take and return functions of the same type, 
and arguments must satisfy one of the input types of an applied function.

Since both of these `referent set composition' operations at the conjunctive nodes -- 
as well as the union operation at the disjunctive nodes -- are linear in space and time 
on the number of elements in each of the composed sets (assuming the sets are 
sorted in advance and remain so), the calculation of referent sets only adds a 
factor of $|\mathcal{E}|$ to the size complexity of the forest and the time 
complexity of processing it, where $|\mathcal{E}|$ is the number of objects 
and events in the run-time environment.
Thus, the total space and time complexity of the above algorithm (on a context-free 
forest) is \order{|\mathcal{E}| n^3}.
If other operations are added, the complexity of referent set composition will 
be limited by the least efficient operation.

\subsection{Temporal referents}

Since the referent sets for situations are also well defined under type theory, 
this environment-based approach can also resolve attachment ambiguities involving 
verbs and verb phrases in addition to those involving only nominal referents.
For example, if the interpreter is given the sentence ``Coolant drained after 
test at 3:00,'' which could mean the draining was at 3:00 or the test was 
at 3:00, the referents for the draining process and the testing process can 
be treated as time intervals in the environment history.%
\footnote{The composition of time intervals, as well as spatial regions 
and other types of situational referents, is more complex than that outlined 
for objects, but space does not permit a complete explanation.}
First, a forest is constructed which shares the subtrees for ``the test'' and 
``after 3:00,'' and the corresponding sets of referents.
Each node in this forest (shown in Figure~\ref{fig:testforest}) is 
then annotated with the set of objects and intervals that it could refer 
to in the environment.
Since there were no testing intervals at 3:00 in the environment, the referent set for 
the NP `test after 3:00' is evaluated to the null set.
But since there is an interval corresponding to a draining process ($\sigma_1$) at the root, 
the whole VP will still be preferred as constituent due to the other interpretation.

\subsection{Quantifier scoping}

The evaluation of referents for quantifiers also presents a tractability problem, 
because the functions they correspond to in the Montague analysis map two sets of 
entities to a truth value.
This means that a straightforward representation of the potential referents 
of a quantifier such as `at least one' would contain every pair of non-empty 
subsets of the set $\mathcal{E}$ of all entities, with a cardinality on the 
order of $2^{2|\mathcal{E}|}$.
If the evaluation of referents is deferred until quantifiers are composed with 
the common nouns they quantify over, the input sets would still be as large as 
the power sets of the nouns' potential referents.
Only if the evaluation of referents is deferred until complete NPs are 
composed as arguments (as subjects or objects of verbs, for example) 
can the output sets be restricted to a tractable size.

This provision only covers {\em in situ} quantifier scopings, however.
In order to model raised scopings, arbitrarily long chains of raised 
quantifiers (if there are more than one) would have to be evaluated 
before they are attached to the verb, as they are in a CCG-style function 
composition analysis of raising \cite{park96}.%
\footnote{This approach is in some sense wedded to a CCG-style syntacto-semantic 
analysis of quantifier raising, inasmuch as its syntactic and semantic structures 
must be isomorphic in order to preserve the polynomial complexity of the shared forest.}
Fortunately, universal quantifiers like `each' and `every' only choose the 
one maximal set of referents out of all the possible subsets in the power set, 
so any number of raised universal quantifier functions can be composed into a 
single function whose referent set would be no larger than the set of all 
possible entities.

It may not be possible to evaluate the potential referents of 
non-universal raised quantifiers in polynomial time, because the number of 
potential subsets they take as input is on the order of the power set of 
the noun's potential referents.
This apparent failure may hold some explanatory power, however, 
since raised quantifiers other than `each' and `every' seem to 
be exceedingly rare in the data.
This scarcity may be a result of the significant computational complexity of 
evaluating them in isolation (before they are composed with a verb).


\section{Evaluation}

An implemented system incorporating this environment-based approach to 
disambiguation has been tested on a set of manufacturer-supplied 
aircraft maintenance instructions, using a computer-aided design (CAD) 
model of a portion of the aircraft as the environment.
It contains several hundred three dimensional objects (buttons, handles, 
sliding couplings, etc), labeled with object type keywords and connected to other 
objects through joints with varying degrees of freedom (indicating how each 
object can be rotated and translated with respect to other objects in the 
environment).


The test sentences were the manufacturer's instructions 
for replacing a piece of equipment in this environment.
The baseline grammar was not altered to fit the test sentences or 
the environment, but the labeled objects in the CAD model 
were automatically added to the lexicon as common nouns.

In this preliminary accuracy test, forest nodes that correspond to noun phrase 
or modifier categories are dispreferred if they have no potential entity referents, 
and forest nodes corresponding to other categories are dispreferred if their 
arguments have no potential entity referents.
Many of the nodes in the forest correspond to noun-noun modifications, 
which cannot be ruled out by the grammar because 
the composition operation that generates them seems to be productive 
(virtually any `N2' that is attached to or contained in an `N1' can be an `N1 N2').
Potential referents for noun-noun modifications are calculated by a rudimentary 
spatial proximity threshold, such that any potential referent of the {\em modified} 
noun lying within the threshold distance of a potential referent of the 
{\em modifier} noun in the environment is added to the composed set.

The results are shown below.
The average number of parse trees per sentence in this set was $21$ 
before disambiguation.
The average ratio of nodes in enumerated tree sets to nodes in shared 
forests for the instructions in this test set was $9.6 : 1$, 
a nearly tenfold reduction due to sharing.

Gold standard `correct' trees were annotated by hand using the 
same grammar that the parser uses.
The success rate of the parser in this domain (the rate at which 
the correct tree could be found in the parse forest) was $87\%$.
The retention rate of the environment-based filtering mechanism 
described above (the rate at which the correct tree was retained 
in parse forest) was $92\%$ of successfully parsed sentences.
The average reduction in number of possible parse trees due to the 
environment-based filtering mechanism described above was $3.6 : 1$ 
for successfully parsed and filtered forests.%
\footnote{Sample parse forests and other details of this 
application and environment are available at 
{\tt http://www.cis.upenn.edu/$\sim$schuler/ebd.html}.}

\small
\begin{center}
\begin{tabular}{c|c|c|c|c}
     & \# trees & nodes in & nodes in & \# trees \\
sent & (before  & unshared & shared   & (after   \\
no.  & filter)  & tree set & forest   & filter)  \\
\hline
 1 &   39 &   600 &  55 &  6 \\
 2 &    2 &    22 &  14 &  2 \\
 3 &   14 &   233 &  32 & 14 \\
 4 &   16 &   206 &  40 &  1 \\
 5 &  36* &   885 &  45 &  3** \\
 6 &   10 &   136 &  35 &  1 \\
 7 &   17 &   378 &  49 &  4 \\
 8 &   23 &   260 &  35 &  3 \\
 9 &   32 &   473 &  35 &  0** \\
10 &   12 &   174 &  34 &  2 \\
11 &  36* &   885 &  45 &  3** \\
12 &   19 &   259 &  37 &  2 \\
13 &    2 &    22 &  14 &  2 \\
14 &   14 &   233 &  32 & 14 \\
15 &   39 &   600 &  55 &  6 \\
\end{tabular}
\end{center}

\ \ \ * indicates correct tree not in parse forest

\ \ \ ** indicates correct tree not in filtered forest

\normalsize



\section{Conclusion}

This paper has described a method by which the 
potential environment referents for all possible interpretations of 
of an input sentence can be evaluated {\em during} parsing, in polynomial time.
The architecture described in this paper has been 
implemented with a large coverage grammar as a run-time interface to 
a virtual human simulation.
It demonstrates that a natural language interface architecture 
that uses the objects and events in an application's run-time 
environment to inform disambiguation decisions (by performing semantic 
evaluation during parsing) is feasible for interactive applications.


\small



\end{document}